\documentclass[11pt]{article}
\usepackage[utf8]{inputenc}
\usepackage{amsmath,amsthm}
\usepackage{amsfonts}
\usepackage{amssymb}
\usepackage{array}
\usepackage{bm}
\usepackage{multirow}
\usepackage{subfig}
\usepackage{enumitem}
\usepackage[normalem]{ulem}
\usepackage[font=small]{caption}
\usepackage{graphicx}

\usepackage[margin=2.9cm]{geometry}
\RequirePackage[colorlinks,citecolor=blue,urlcolor=blue]{hyperref}

\usepackage{algorithm}
\usepackage{algpseudocode}

\algnewcommand{\IIf}[1]{\State\algorithmicif\ #1\ \algorithmicthen}

\usepackage[%
natbib=true,
style=numeric,
sorting=none,
backend=bibtex,
doi=false,
isbn=false,
url=false,
giveninits=true,
maxnames=4
]{biblatex}

\makeatletter
\def\blx@maxline{77}
\makeatother

\renewbibmacro{in:}{}           
\DeclareFieldFormat[article,inbook,incollection,inproceedings,patent,thesis,unpublished]{citetitle}{#1}
\DeclareFieldFormat[article,inbook,incollection,inproceedings,patent,thesis,unpublished]{title}{#1} 

\usepackage{arash_macros}
\usepackage{authblk}
\title{Spectrally-truncated kernel ridge regression and its free lunch}

\author{%
	Arash A. Amini%
	\\
	Department of Statistics\\
	University of California, Los Angeles\\
}

\newcommand{\us}{u^*}

\newcommand{\fs}{f^*}
\newcommand{\foms}{f_{\omega^*}}
\newcommand{\ft}{\widetilde f}
\newcommand{\Hil}{\mathcal H}

\newcommand{\omt}{\widetilde \omega}
\newcommand{\oms}{\omega^*}
\newcommand{\Ker}{\mathbb K}
\newcommand{\Kerarg}[1]{\Ker(\,\cdot\,,#1)}
\newcommand{\Lc}{\mathcal{L}}
\newcommand{\yt}{\widetilde{y}}

\newcommand{\eerr}{\text{EE}_{r,\lambda}}
\newcommand{\ee}{\text{EE}}
\newcommand{\wae}{\text{WAE}}

\newcommand{\waerr}{\text{WAE}_{r,\lambda}}

\DeclareMathOperator{\mse}{MSE}

\newcommand{\Xc}{\mathcal X}

\newcommand{\hip}[1]{\ip{#1}_\Hil}
\newcommand{\empnorm}[1]{\norm{#1}_n}

	\newcommand{\wt}{\widetilde{w}}
	
\newcommand{\vs}{v^*}

	\newcommand{\overbar}{\bar}

\newcommand{\omb}{\overbar{\omega}}

\newcommand{\ub}{\overbar{u}}

\newcommand{\fb}{\overbar{f}}

\newcommand{\Psil}{\Psi_\lambda}

\newcommand{\Kt}{\widetilde K}
\newcommand{\Dt}{\widetilde D}
\newcommand\Kap{\Kt}
\newcommand\Dap{\Dt}
\newcommand{\Gaml}{\Gamma_\lambda}
\newcommand{\wtt}{\widehat w}
\DeclareMathOperator{\cov}{cov}
\newcommand\hilnorm[1]{\norm{#1}_\Hil}

\begin{document}
	\maketitle
	\begin{abstract}
		Kernel ridge regression (KRR) is a well-known and popular nonparametric regression approach with many desirable properties, including minimax rate-optimality in estimating functions that belong to common reproducing kernel Hilbert spaces (RKHS). The approach, however, is computationally intensive for large data sets, due to the need to operate on a dense $n \times n$ kernel matrix, where $n$ is the sample size. Recently, various approximation schemes for solving KRR have been considered, and some analyzed. Some approaches such as Nystr\"{o}m approximation and sketching have been shown to preserve the rate optimality of KRR. In this paper, we consider the simplest approximation, namely, spectrally truncating the kernel matrix to its largest $r < n$ eigenvalues. We derive an exact expression for the maximum risk of this truncated KRR, over the unit ball of the RKHS. This result can be used to study the exact trade-off between the level of spectral truncation and the regularization parameter. We show that, as long as the RKHS is infinite-dimensional, there is a threshold on $r$, above which, the spectrally-truncated KRR surprisingly outperforms the full KRR in terms of the minimax risk, where the minimum is taken over the regularization parameter. This strengthens the existing results on approximation schemes, by showing that not only one does not lose in terms of the rates, truncation can in fact improve the performance, for all finite samples (above the threshold). Moreover, we show that the implicit regularization achieved by spectral truncation is not a substitute for Hilbert norm regularization. Both are needed to achieve the best performance.

	\end{abstract}

\maketitle

\section{Introduction}

The general nonparametric regression problem %
can be stated as 
\begin{align}\label{eq:mod:1}
y_i = \fs(x_i) + w_i, \;i = 1,\dots,n, \quad \ex w = 0,\; \cov(w) = \sigma^2I_n
\end{align}
where $w = (w_i) \in \reals^n$ is a noise vector and $\fs: \Xc \to \reals$ is the function of interest to be approximated from the noisy observations $\{y_i\}$. Here, $\Xc$ is the space to which the \emph{covariates} $\{x_i\}$ belong. We consider the fixed design regression where the covariates are assumed to be deterministic.
The problem has a long history in statistics and machine learning~\citep{wasserman2006all,tsybakov2009intro}. In this paper, we assume that $\fs$ belongs to a reproducing kernel Hilbert space (RKHS), denoted as $\Hil$~\citep{paulsen2016introduction}. Such spaces are characterized by the existence of a reproducing kernel, that is, a positive semidefinite function $\Ker : \Xc \times \Xc \to \reals$ that uniquely determines the underlying function space $\Hil$. RKHSs are very versatile modeling tools and  include, for example, Sobolev spaces of smooth functions whose norms are measures of function roughness%
~\citep{wahba1990spline}. Throughout, we think of these Sobolev spaces as the concrete examples of $\Hil$. By assuming an upper bound on the Hilbert norm of $\fs$, we can encode a prior belief that the true data generating function $\fs$ has a certain degree of smoothness. Without loss of generality, we assume that $\fs$ belongs to the unit ball of the RKHS, that is,
\begin{align}\label{assu:unit:ball:f}
\fs \;\in\; \ball_\Hil := \{f \in \Hil:\; \norm{f}_\Hil \le 1\}.
\end{align}
A natural estimator is then, the kernel ridge regression (KRR), defined as the solution of the following optimization problem:
\begin{align}\label{eq:krr:f:1}
\ft_{n,\lambda} := \min_{f \in \Hil} 
\;\frac1{n} \sum_{i=1}^n (y_i - f(x_i))^2 + \lambda \norm{f}_\Hil^2,
\end{align}
where $\lambda > 0$ is a regularization parameter. It is well-known that this problem can be reduced to a finite-dimensional problem, by an application of the so-called representer theorem~\citep{kimeldorf1971some}:
\begin{align}\label{eq:krr:omg:1}
\min_{\omega \,\in\, \reals^n} \; 
\frac1{n} \vnorm{y- \sqrt n K \omega}^2 + \lambda
\omega^T K \omega, \quad \text{where}\quad  K =\frac1n \big( \Ker(x_i,x_j)\big) \in \reals^{n \times n}
\end{align} 
is the (normalized empirical) kernel matrix. Although~\eqref{eq:krr:omg:1} has a closed form solution, it involves inverting an $n \times n$ dense matrix, with time complexity $O(n^3)$,
which is prohibitive in practice. 

Various approximation schemes have been proposed to mitigate the computational costs, including (i) approximating the kernel matrix or  (ii) directly approximating the optimization problem~\eqref{eq:krr:omg:1}. Examples of the former are the Nystr\"{o}m approximation, column sampling and their variants~\citep{williams2001using,zhang2008improved,kumar2009ensemble,li2010making,talwalkar2014matrix}. An example of the latter is sketching~\citep{alaoui2015fast,yang2017randomized} where one restricts $\omega$ to the subspace $\ran(S) := \{S\alpha \mid  \alpha \in\reals^r\}$, for some random matrix $S \in \reals^{n \times r}$.   It is in fact known that Nystr\"{o}m can be considered a special case of sketching with random standard basis vectors~\citep{yang2017randomized}.
Sketching, with sufficiently large $r$, has been shown in~\cite{yang2017randomized} to achieve minimax optimal rates over Sobolev spaces, under mild conditions on the sketching matrix $S$. Similarly, the Nystr\"{o}m approximation has been analyzed in~\cite{cortes2010impact,yang2012nystrom,jin2013improved,alaoui2015fast,bach2013sharp} and~\cite{rudi2015less},   the latter showing minimax rate optimality.
In addition to the above, (iii) divide and conquer approaches have been proposed~\citep{zhang2013divide}, where one solves the problem over subsamples and then aggregates by averaging, with some rate optimality guarantees. Other notable approaches to scaling include (iv) approximating translation-invariant kernel functions via Monte Carlo averages of tensor products of randomized feature maps~\citep{rahimi2008random,le2013fastfood} %
and (v) applying stochastic gradient in the function space~\citep{dai2014scalable}. Memory efficiency in kernel approximation is considered in~\cite{si2017memory}.

In this paper, we consider the most direct kernel approximation, namely, replacing $K$ by its best rank $r$ approximation (in Frobenius norm). This amounts to truncating the eigenvalue decomposition of $K$ to its top $r$ eigenvalues. We refer to the resulting KRR approximation as the \emph{spectrally-truncated} KRR (ST-KRR). Although somewhat slower than the Nystr\"{o}m approximation and fast forms of sketching, ST-KRR can be considered an ideal rank-$r$ spectral approximation. By analyzing it, one can also gain insights about approximate SVD truncation approaches such as Nystr\"{o}m or sketching. Practically, ST-KRR is a very viable solution for moderate-size problems. See Appendix~\ref{app:time} for a discussion of the time complexity of various schemes.

We derive an exact expression for the maximum (empirical) mean-squared error (MSE) of ST-KRR, uniformly over the unit ball of the RKHS. This expression is solely in terms of the eigenvalues $\{\mu_i\}$ of the kernel matrix $K$, the regularization parameter $\lambda$, the truncation level $r$,  and the noise level $\sigma^2$. Thus if one has access to $\{\mu_i\}$ and the noise level (or estimates of them), one can plot the exact regularization curve (maximum MSE versus $\lambda$) for a given truncation level $r$ and sample size $n$, and determine the  optimal value of $\lambda$. %
We also note that since the empirical eigenvalues $\{\mu_i\}$ quickly approach those of the integral operator associated with $\Ker$, as $n \to \infty$~\citep{koltchinskii2000random}, one can use these idealized eigenvalues instead of $\{\mu_i\}$ to get an excellent approximation of these regularization curves. 

We then show that there is an optimal threshold on $r$, the truncation level, which we denote as $r_n$, such that for all $r \ge r_n$, the minimax risk of the $r$-truncated KRR, with the minimum taken over the regularization parameter, is strictly smaller than that of the full KRR whenever $\mu_{r+1} > 0$. For infinite-dimensional RKHSs, we always have $\mu_{r_n+1} > 0$, hence truncating at level $r_n$ is guaranteed to strictly improve performance.  The slower the decay of the eigenvalues, the larger this gap in performance.

This result shows that although the spectral truncation is mainly used as a computational device, it also has a statistical regularization effect. The next question is whether the regularization provided by the spectral truncation renders Hilbert norm regularization (via $\lambda$) unnecessary. We answer this question in the negative by showing that for any truncation level $r$, the optimal maximum risk is achieved for a positive $\lambda$. Together, these results show that the ``$r$-truncated $\lambda$-regularized KRR'' defines a new class of estimators whose performance cannot be achieved (in finite sample) with either regularization alone.

We also show how the exact expression for the maximum MSE can be used to easily establish a slightly weaker bound for ST-KRR, similar to those derived in~\cite{yang2017randomized} for sketching. We  discuss the link between the statistical dimension considered in~\cite{yang2017randomized} and the optimal truncation level $r_n$, and show how the same rate-optimality guarantees hold for ST-KRR. Rate-optimality also follows form the fact that ST-KRR, with proper $r$, strictly dominates full KRR and the latter is rate-optional. However, we do these calculations to make the comparison easier. 

Finally, we illustrate the results with some numerical simulations showing some further surprises. For example, the Gaussian kernel has a much faster eigendecay rate than a Sobolev-1 kernel (exponential versus polynomial decay).  Hence, the optimal truncation level $r_n$ asymptotically grows much slower %
for the Gaussian kernel. However, for finite samples, depending on the choice of the Gaussian bandwidth, the exact optimal truncation level, computed numerically, can be larger than that of Sobolev-1.

\section{Preliminaries}
Let us start with some observations regarding the original KRR problem in~\eqref{eq:krr:f:1}.
For $\omega \in \reals^n$, consider the \emph{kernel mapping} %
\begin{align}\label{eq:f:om}
f_\omega := \frac1{\sqrt n} \sum_j \omega_j \Ker(\cdot,x_j).
\end{align}
Note that $\omega \mapsto f_\omega$ is a linear map from $\reals^n \to \Hil$.  This map is the link between the solutions of the two optimization problems~\eqref{eq:krr:f:1} and~\eqref{eq:krr:omg:1}: For any optimal solution $\omega$ of~\eqref{eq:krr:omg:1}, $f_\omega$ will be an optimal solution of~\eqref{eq:krr:f:1}. The link is easy to establish by observing the following two identities:
\begin{align}\label{eq:hilnorm:ident}
\norm{f_\omega}_\Hil^2 = \omega^T K \omega, \quad f_\omega(x_i) = \sqrt n (K \omega)_i,
\end{align}
the first of which uses the \emph{reproducing property} of the kernel: $\hip{f, \Kerarg{x}} =f(x)$. We will frequently use this property in the sequel.
The proof of the equivalence follows from an argument similar to our discussion of the identifiability below.

\subsection{Identifiability}\label{sec:indentify}
Let us first observe that $\fs$ in~\eqref{eq:mod:1} is not (statistically) identifiable. That is, there are multiple functions $\fs$ (in fact, infinitely many if $\Hil$ is infinite-dimensional) for which the vector $(y_i)$ has the exact same distribution. To see this, let
\begin{align}\label{eq:LcX}
\Lc_X := \text{span}\{ \Kerarg{x_i}:\; i \in [n]\} = \{f_\omega :\; \omega \in \reals^n\},
\end{align}
and let $f_{\oms}$ be the projection of $\fs$ onto $\Lc_X$. (It is always possible to choose at least one such $\oms$ by the definition of projection and since $\Lc_X$ is a closed subspace of $\Hil$.)
Given observations $(y_i)$, we can only hope to recover the following equivalence class:
\begin{align*}
\{ f \in \Hil :\; f(x_i) = \fs(x_i), \; \forall i\} &= \{ f \in \Hil :\; \ip{f-\fs, \Kerarg{x_i}}_\Hil = 0, \; \forall i\} \\
& = \{ \fs + g:\; g \perp_\Hil \Lc_X\} \\
&= \{ \foms + g:\; g \in \Lc_X^\perp\} = \foms + \Lc_X^\perp
\end{align*}
where the last line follows since $\fs - \foms \in \Lc_X^\perp$ by the property of orthogonal projection (and can be absorbed into $g$).

We will use $\foms$ as the representative of the (identifiable) equivalence class of $\fs$.
We are interested in measuring functional deviations (e.g., the error in our estimate relative to the true function) in the \emph{empirical $\ell_2$ norm}:
\begin{align*}
\empnorm{f-g} = \Big[\frac1n \sum_{i=1}^n \big(f(x_i) - g(x_i) \big)^2\Big]^{1/2}.
\end{align*}
The use of this norm is %
common in the literature of nonparametric regression~\citep{van2000empirical,wainwright2019high}. It is interesting to note that $\vnorm{\fs-\foms}_n = 0$, 
\begin{align}\label{eq:equiv:empnorm}
\empnorm{f-\fs} = \empnorm{f-\foms}, \quad \forall f \in \Hil,
\end{align}
and $\vnorm{\foms}_\Hil \le \vnorm{\fs}_\Hil$, since projections are contractive.
Thus, recalling~\eqref{assu:unit:ball:f}, $\foms$ also belongs to the Hilbert unit ball: $\foms \in \ball_\Hil$. It is in fact easy to see that $\foms$ has the least Hilbert norm among the members in the equivalence class (i.e., the smoothest version). Thus, without loss of generality, we can identify $\fs$ with $\foms$. Equivalently, we can assume from the start that $\fs$ is of the form $\foms$ for some $\oms \in \reals^n$. Note that the ``no loss of generality'' statement holds as long as we are working with the empirical $\ell_2$ norm, due to~\eqref{eq:equiv:empnorm}.

\section{Main results}

Let $K = U DU^T$ be the eigenvalue decomposition (EVD) of the empirical kernel matrix defined in~\eqref{eq:krr:omg:1}. Here, $U \in \reals^{n \times n}$ is an orthogonal matrix and $D = \diag(\mu_i)_{i=1}^n$ where $\mu_1 \ge \mu_2 \ge \dots \ge \mu_n \ge 0$ are the eigenvalues of $K$. We assume for simplicity that $\mu_n> 0$, that is, the exact kernel matrix is invertible. Consider the rank $r$ approximation of $K$, obtained by keeping the top $r$ eigenvalues and truncating the rest to zero, that is,
\begin{align*}
\Kap = \Kap_r :=  U
\begin{pmatrix}
D_r & 0 \\
0 & 0
\end{pmatrix} U^T = U_r D_r U_r^T.
\end{align*}
Here, $D_r = \diag(\mu_1,\dots,\mu_r)$ and $U_r \in \reals^{n \times r}$ collects the first $r$ columns of $U$. 
The idea is to solve~\eqref{eq:krr:omg:1} with $K$ replaced with $\Kap$, to obtain $\omt$. We then form our functional estimate $\ft$ by using the (exact) kernel mapping~\eqref{eq:f:om}. 
\begin{definition}\label{dfn:trunc:krr}
	An \emph{$r$-truncated $\lambda$-regularized KRR estimator} with input $y \in \reals^n$ , is a function $\ft := f_{\omt} = \frac1{\sqrt n} \sum_j \omt_j \Ker(\cdot,x_j)$ where
	\vspace{-1.5ex}
	\begin{align}
	\omt \in &\argmin_{\omega \,\in\, \reals^n} \label{eq:omt:min}  \; 
	\frac1{n} \vnorm{y- \sqrt n \Kap \omega}^2 + \lambda \omega^T \Kap \omega, \\
	&\text{such that }  \Kap \omt = K \omt\label{eq:omt:cond}.
	\end{align}
\end{definition}

A minimizer in~\eqref{eq:omt:min}, without the additional condition $\Kap \omt = K \omt$, is not unique due to the rank deficiency of $\Kap$. Thus, we can ask for it to satisfy additional constraints. The equality condition in~\eqref{eq:omt:cond}, which can be stated as $\omt \in \ker(\Kap - K)$ can always be satisfied. It is enough to choose $\omt$ to be the unique minimizer in $\ran(\Kap) = \ran(U_r)$, that is, $\omt = U_r \alpha$ for some $\alpha \in \reals^r$. This  is how the estimator is often implemented in practice.

\smallskip
We are interested in the deviation of $\ft$ from the true function $\fs$ in the empirical $\ell_2$ norm. More precisely, we are interested in the mean-squared error as the statistical risk:
\begin{align*}
\mse(\ft,\fs) = \ex \empnorm{\ft - \fs}^2.
\end{align*}
Our main result is an expression for the worst-case risk of $\ft$ over the unit ball of the RKHS:
\begin{theorem}\label{thm:krr:spec:trunc}		
	Let $\ft = \ft_{r,\lambda}$ be an $r$-truncated $\lambda$-regularized KRR estimator (Definition~\ref{dfn:trunc:krr}) applied to input $y$ generated from model~\eqref{eq:mod:1}. Let
	\begin{align*}
	H_r(\lambda) := \max_{1 \le i \le r} h(\lambda;\mu_i)
	\end{align*}
	where  $h(\lambda;x) = \lambda^2 x/(x+\lambda)^2$.
	Then, for all $r=1,2,\dots,n$ and $\lambda > 0$,
	\begin{align}\label{eq:krr:trunc:res}
	\sup_{\fs \, \in \, \ball_\Hil} \mse(\ft_{r,\lambda},\fs) 
	\;=\; \max\big\{ 
	H_r(\lambda), \;
	\mu_{r+1} \big\} + \frac{\sigma^2}{n} \sum_{i=1}^r \Big(\frac{\mu_i}{\mu_i + \lambda}\Big)^2,
	\end{align} 
	with $\mu_{n+1} := 0$.
\end{theorem}

The first term in~\eqref{eq:krr:trunc:res} is the worst-case approximation error (WAE) and the second term the estimation error (EE). The approximation error (AE) is the risk (relative to $\fs$) of $\fb$  which is obtained by passing the noiseless observations $(\fs(x_i))$, instead of $y$, through the estimation procedure. The AE is the deterministic part of the risk and is given by $\empnorm{\fb-\fs}^2$. The estimation error is the stochastic part of the risk and is given by $\ex \empnorm{\ft - \fb}^2$.

\medskip
The function $x \mapsto h(\lambda;x)$ attains its maximum of $\lambda/4$, over $[0,\infty)$, at $x = \lambda$. Thus, as long as $\lambda \in [\mu_r, \mu_1]$,
the bound $H_r(\lambda) \le \lambda/4$ is  good. In general,
\begin{align}\label{eq:wae:bound}
\waerr \le \max\Big\{\frac\lambda4 , \;\mu_{r+1} \Big\}.
\end{align} 
We note that since the KRR estimates are linear in $y$, Theorem~\ref{thm:krr:spec:trunc} easily gives the maximum MSE expression over the Hilbert ball of arbitrary radius $R$, by replacing $\sigma^2$ in~\eqref{eq:krr:trunc:res} with $\sigma^2/R^2$ and multiplying the entire right-hand side by $R^2$.

We also have a precise result on the regularized risk of the approximating function:

\begin{proposition}\label{lem:approxerr:pluse:smoothness:good}
	Let  $\fb = \fb_{r,\lambda}$ be obtained by passing the noiseless observations $(\fs(x_i))$, instead of $y$, through the estimation procedure in Definition~\ref{dfn:trunc:krr}. Then,
	\begin{align}\label{eq:approxerr:pluse:smoothness:good}
	\sup_{\fs \,\in\, \ball_\Hil}	\vnorm{\fs - \fb}_n^2 + \lambda \vnorm{\fb}_\Hil^2 
	\;=\; \max\Big\{\max_{1\le i \le n} \frac{\lambda \mu_i}{\mu_i + \lambda}, \;\mu_{r+1}\Big\}.
	\end{align}
\end{proposition}

\subsection{Maximum-risk inadmissibility}
Let us now consider how the maximum  risk of the truncated  KKR compares with the full version. For every, $\lambda > 0$, define
\begin{align*}
r(\lambda) := \min\{r \in [n]:\; \mu_{r+1} \le H_n(\lambda) \}.
\end{align*}
In addition, recalling that $\ft_{n,\lambda}$ is the full KRR estimator, let 
\begin{align}\label{eq:lambda:n}
\lambda_n := \argmin_{\lambda > 0} \sup_{\fs \, \in \, \ball_\Hil} \mse(\ft_{n,\lambda},\fs), \quad \text{and} \quad 
r_n := r(\lambda_n).
\end{align}

That is, $\lambda_n$ is the regularization parameter that achieves the minimal maximum-risk for the full KRR.
We have the following corollary of Theorem~\ref{thm:krr:spec:trunc}:
\begin{corollary}\label{cor:domination}
	For every $\lambda > 0$, and every $r \in [n]$ with $r \ge r(\lambda)$,
	\begin{align}\label{eq:max:risk:domination}
	\sup_{\fs \, \in \, \ball_\Hil} \mse(\ft_{r,\lambda},\fs) 
	\;\le\;
	\sup_{\fs \, \in \, \ball_\Hil} \mse(\ft_{n,\lambda},\fs).
	\end{align}
	In particular, for every $r \ge r_n$,
	\begin{align}\label{eq:minmax:risk:domination}
	\min_{\lambda > 0} \sup_{\fs \, \in \, \ball_\Hil} \mse(\ft_{r,\lambda},\fs) 
	\;\le \;
	\min_{\lambda > 0} \sup_{\fs \, \in \, \ball_\Hil} \mse(\ft_{n,\lambda},\fs).
	\end{align}
	Both inequalities are strict whenever $\mu_{r+1} > 0$. %
\end{corollary}

Corollary~\ref{cor:domination} shows that $\lambda$-optimized $\ft_{r_n,\lambda}$ strictly improves on optimized full KRR whenever $\mu_{r_n+1} > 0$, in a sense rendering the full KRR inadmissible, as far as the maximum risk over $\ball_\Hil$ is concerned. Note that we are not claiming inadmissibility in the classical sense which requires one estimator to improve on another for all $\fs \in \ball_\Hil$.  In general, the slower the decay of $\{\mu_i\}$, the more significant the improvement gained by truncation. Note that~\eqref{eq:lambda:n} allows one to set the precise truncation level including the exact constants if one has access to the eigenvalues of the kernel matrix. In practice, for large $n$, the eigenvalues of the associated kernel integral operator (if available) can act as excellent surrogates for $\{\mu_i\}$~\citep{koltchinskii2000random}.

\subsection{Do we need both regularizations?}
Although the spectral truncation is used as a computational device, intuitively, it also has an implicit regularization effect. This is confirmed more rigorously by Corollary~\ref{cor:domination} where truncation is shown to lead to a smaller optimal worst-case MSE. The intuition is also supported by the link between the (full) KRR and  Tikhonov regularization. In both cases, one forms $(K+ \lambda I_n)^{-1}$ which can be considered as a form of ``spectral filtering''. Eigenvalue truncation followed by taking the pseudo-inverse can be considered as another form of such filtering. A common conception is that these two approaches are performing essentially the same task, hence one of them is enough to achieve the desired regularization effect. More specifically, one can ask the following: Is Hilbert norm regularization, or $\lambda$-regularization, really needed in the presence of spectral truncation? Theorem~\ref{thm:krr:spec:trunc} allows us to settle this question. For a given truncation level~$r$, let
\begin{align}\label{eq:lambda:r:def}
\lambda_r := \argmin_{\lambda > 0} \sup_{\fs \, \in \, \ball_\Hil} \mse(\ft_{r,\lambda},\fs)
\end{align}
be the optimal threshold for the $r$-truncated $\lambda$-regularized KRR estimator.

\begin{corollary}\label{cor:lambda:r}
	For every $r < n$, we have
	\begin{align*}
	\lambda_r \ge \max \Big\{ \frac{\mu_r}{\sqrt{\mu_r/\mu_{r+1}} - 1}, \frac{\sigma^2}n \big(1+ B_r\big) \Big\}
	\end{align*}
	where
	$
	B_r :=  \min_{j} \sum_{i:\,i > j} (\mu_{j}/ \mu_i)  + \sum_{i: \,i < j} (\mu_i^2/ \mu_{j}^2)
	$
	with $i$ and $j$ running in $\{1,\dots,r\}$.
\end{corollary}
Corollary~\ref{cor:lambda:r} shows that for any truncation level $r$, the optimal choice of $\lambda$ is always positive, hence $\lambda$-regularization further improves the performance. The effect is more pronounced when $\mu_r$ is close to $\mu_{r+1}$ or, in general, when the spectrum decays slowly (hence $\mu_i \approx \mu_j$ for most $i,j \in [r]$). The effect is also more significant for  higher effective noise levels $\sigma^2/n$.

\subsection{Gaussian complexity and rates}%
Less precise bounds, albeit good enough to capture the correct asymptotic rate as $n \to \infty$, can be obtained in terms of the Gaussian complexity of the unit ball of the RKHS. These types of results have been obtained for the Sketched-KRR. To make a comparison easier, let us show how such bounds can be obtained from Theorem~\ref{thm:krr:spec:trunc}.

Let us define the $r$-truncated complexity (of the empirical Hilbert ball) as
\begin{align}\label{eq:Gauss:complexity}
R_r(\delta) =  \Big( \frac{\sigma^2}{n}\sum_{i=1}^r \min\{\mu_i, \delta^2\}\Big)^{1/2}.
\end{align}
For the case $r=n$, this matches the definition of the kernel complexity in~\citep{yang2017randomized}, which we refer to for the related background. In particular,~\eqref{eq:Gauss:complexity} is a tight upper bound on the Gaussian complexity of the intersection of $\ball_\Hil$ and $\{f:\;\norm{f}_n \le \delta \}$~\citep[Chapter 13]{wainwright2019high}. We have:
\begin{corollary}[Looser bound]\label{cor:Gauss}
	Under the setup of Theorem~\ref{thm:krr:spec:trunc}, for $\lambda \ge \max\{\delta^2, 4\mu_{r+1}\}$,
	\begin{align}\label{eq:weaker:upper:bound}
	\sup_{\fs \,\in\, \ball_\Hil} \mse(\ft_{r,\lambda}, \fs) \;\le\; \frac14 \lambda + \Big(\frac{R_r(\delta)}{\delta} \Big)^2.
	\end{align}
	If $\lambda \ge \mu_1$, one can replace the first term with $\mu_1\lambda^2/(\lambda+\mu_1)^2$ for a better bound.
\end{corollary}

Choosing $\lambda = \delta^2 \ge 4 \mu_{r+1}$, we obtain
\[
\text{RMSE} = \sqrt{\mse} \le \frac{\delta}2 +  \frac{R_r(\delta)}{\delta} \le \frac{\delta}2 +  \frac{R_n(\delta)}{\delta}.
\]
The latter upper bound is what one would get for the full KRR. 
Matching the two terms in that bound, we chooses $\delta_n$ such that $ \delta_n^2 = 2 R_n(\delta_n)$ which gives the well-known critical radius for the KRR problem~\citep{wainwright2019high}. It is known that 
$\delta_n$ gives the optimal rate of convergence for estimating functions in $\ball_\Hil$, i.e., its rate of decay matches that of the minimax risk~\citep{yang2017randomized}. The above argument shows that as long as $r$ is taken large enough so that $4 \mu_{r+1} \le \delta_n^2$,  the $r$-truncated KRR achieves (at least) the same rate as the full KRR. For the sketching, the same conclusion is established in~\cite{yang2017randomized}, where the smallest $r$ satisfying  $\mu_{r} \le \delta_n^2$ is referred to as the \emph{statistical dimension} of the kernel.

For Sobolev-$\alpha$ kernels, with eigendecay $\mu_i \asymp i^{-2\alpha}$, we obtain $\mse \lesssim \delta_n^2  \asymp (\sigma^2/n)^{-\frac{2\alpha}{2\alpha+1}}$. Interestingly, in this case, the estimate based on the weaker bound~\eqref{eq:weaker:upper:bound} and the exact bound~\eqref{eq:krr:trunc:res} give the same rate (cf.~Appendix~\ref{app:rates}). This is expected since the given rate is known to be minimax optimal for Sobolev spaces. The same goes for the Gaussian kernel for which $\mu_j \asymp e^{- c j \log j}$ and the rate is $\gamma \log(1/\gamma)$ for $\gamma = \sigma^2/n$.

Order-wise, $\delta_n^2$  will be the same as $\lambda_n$ defined in~\eqref{eq:lambda:n},  that is $\lambda_n \asymp \delta_n^2$, whenever $\delta_n^2$ matches the optimal rate. Hence, often $\mu_1 > \lambda_n \gg \mu_n$ for large $n$ and the argument leading to~\eqref{eq:wae:bound} suggests that in this case $H_n(\lambda_n) \approx \lambda_n/4$. Then,
$r_n \approx \min\big\{r \in [n]:\; \mu_{r+1} \le \frac{\lambda_n}{4} \big\}.$

For Sobolev-$\alpha$ kernels, this suggests truncation level $r_n \gtrsim (\sigma^2/n)^{\frac{1}{2\alpha+1}}$ which gives moderate savings for high smoothness levels $\alpha$. Similarly, for the Gaussian kernel, it is not hard to see that truncating to $r_n \gtrsim \log(n/\sigma^2)$ is enough to get the same rate as the full KRR, which is a substantial saving.

\section{Simulations}
\begin{figure}[t]
	\centering

	\includegraphics[width=1\textwidth]{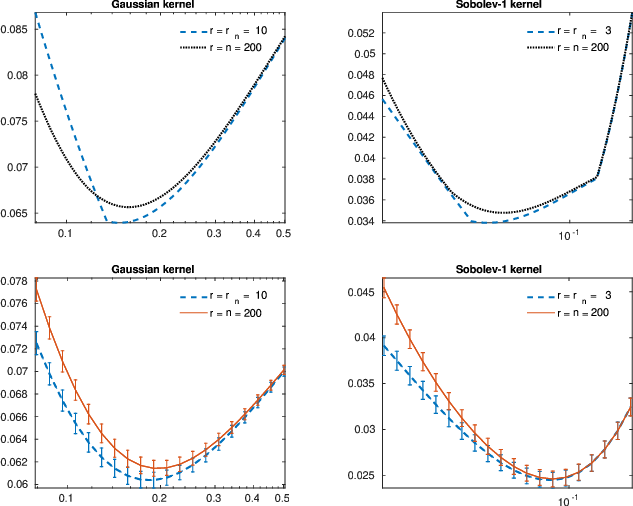} 

	\caption{Plots of (top) the maximum theoretical MSE (bottom) typical empirical MSE, versus the regularization parameter ($\lambda$) for the Gaussian (with bandwidth = 10) and Sobolev-1 kernels on $[-1,1]$ with $n=200$ equispaced samples. The optimally-truncated KRR is shown ($r=r_n$) together with the full KRR ($r=n$). }
	\label{fig:mse}
\end{figure}

We now present some numerical experiments to corroborate the theory. We consider a Gaussian kernel $K(s,t) = e^{-(u-v)^2/2b^2}$ of bandwidth $b = 0.1$ on $[-1,1]$, as well as the Sobolev-1 kernel $K(s,t) = \min(s,t)$ on $[0,1]$. We take the covariates $\{x_i\}$ to be $n=200$ equi-spaced points in each interval. The top row of Fig.~\ref{fig:mse} shows the plot of the theoretical maximum MSE as given by Theorem~\ref{thm:krr:spec:trunc} for the two kernels, for both the full KRR ($r=n$), and the optimally truncated version ($r = r_n$). We have used $\sigma = 2$ in~\eqref{eq:krr:trunc:res}. As predicted by Theorem~\ref{thm:krr:spec:trunc}, the minimum achievable maximum MSE is smaller for the truncated KRR.

To compute the optimal truncation, we have evaluated the regularization curve of the full KRR first, obtained the minimizer $\lambda_n$ and then used~\eqref{eq:lambda:n} to compute the optimal truncation level $r_n$. For the setup of the simulation, we get $r_n = 10$ for the Gaussian and $r_n = 3$ for the Sobolev-1. It is interesting to note that although in terms of rates, $r_n$ for the Gaussian should be asymptotically much smaller than that of Sobolev-1, in finite samples, the truncation level for the Gaussian could be bigger as can be seen here. This is due to the unspecified, potentially large, constants in the rates (that depend on the bandwidth $b$ as well). Also, notice how surprisingly small $r_n$ is relative to $n$ in both cases.

The bottom row of Fig.~\ref{fig:mse} shows the empirical MSE obtained for a typical random $f^* \in \ball_\Hil$, by computing the KRR estimates for observation $y$ and comparing with $\fs$. The random true function is generated as $\fs = f_{\oms}$ where $\oms \sim N(0,I_n)$ and further normalized so that $(\oms)^T K \oms = 1$. We have generated $n=200$ observations from~\eqref{eq:mod:1} with $\sigma = 2$. The plots %
were obtained using 1000 replications. The truncation levels are those calculated based on the maximum MSE formula~\eqref{eq:krr:trunc:res}. The plots show that for a typical application, the truncated KRR also dominates the full KRR. %

\section{Proof of the main result}

Here we give the proof of Theorem~\ref{thm:krr:spec:trunc} and Corollaries~\ref{cor:domination} and~\ref{cor:lambda:r}. The remaining proofs can be found in Appendix~\ref{app:rem:proofs}.

From the discussion in Section~\ref{sec:indentify}, both the KRR estimate and the true function belong to  $\Lc_X$ given in~\eqref{eq:LcX}. It is then useful to have an expression for the empirical $\ell_2$ error of functions belonging to this space. First, we observe that $\vnorm{f_\omega}_n^2 = \frac1n \sum_{i=1}^n [f_\omega(x_i)]^2 = \vnorm{K \omega}^2$. Now, take any $\omega, \oms \in \reals^d$, and let $u = K \omega$ and $\us = K \oms$. Then, we have
\begin{align}\label{eq:emp:uspace:equiv}
\empnorm{f_\omega - \foms} = \empnorm{f_{\omega - \oms}} = \norm{K(\omega - \oms)} = \norm{u - \us}
\end{align}
where the fist equality is by the linearity of $\omega \mapsto f_\omega$.
For any function $f_\omega \in \Lc_X$, we call $u = K\omega$ the \emph{$u$-space} representation of $f_\omega$. Identity~\eqref{eq:emp:uspace:equiv} shows that it is often easier to work in the $u$-space since the \emph{$u$-transform} turns empirical $\ell_2$ norms on functions into the usual $\ell_2$ norms on vectors. In other words, the map $f_\omega \mapsto u$, is a Hilbert space isometry from $(\Lc_X,\empnorm{\cdot})$ to $(\reals^n,\norm{\cdot})$. In the $u$-space, the KRR optimization problem can be equivalently stated as:
\begin{align}\label{eq:krr:u:1}
\min_{u \,\in\, \ran(K)} \; \frac1{n} \vnorm{y- \sqrt{n} u}^2 + \lambda u^T K^{+} u
\end{align}
where $K^+$ is the pseudo inverse of $K$, and $\ran(K)$ its range. More precisely:
\begin{lemma}\label{lem:omega:u:equiv}
	For any $K \in \reals^{n \times n}$, problems~\eqref{eq:krr:omg:1} and~\eqref{eq:krr:u:1} are equivalent in the following sense:
	\begin{enumerate}[label=-]
		\item For any minimizer $\omb$ of~\eqref{eq:krr:omg:1}, $K \omb$ is a minimizer of~\eqref{eq:krr:u:1}, and
		\item  for any minimizer $\ub$ of~\eqref{eq:krr:u:1}, any $\omb \in \{ \omega :\; K \omega = \ub\}$ is a minimizer of~\eqref{eq:krr:omg:1}. 
	\end{enumerate}
\end{lemma}

It is often the case that the kernel matrix itself is invertible, in which case $K^+ = K^{-1}$, $\ran(K) = \reals^n$ and problem~\eqref{eq:krr:u:1} simplifies. However, the equivalence in Lemma~\ref{lem:omega:u:equiv} holds even if we replace $K$ with an approximation which is rank deficient. This observation will be useful in the sequel.

\begin{proof}[Theorem~\ref{thm:krr:spec:trunc}]
	Take $\omt$ to be as in Definition~\ref{dfn:trunc:krr} and let $\yt = y/\sqrt{n}$. Since $\omt$ is the minimizer of
	$
	F(\omega;y) = \vnorm{\yt- \Kap \omega}^2 + \lambda \omega^T \Kap \omega,
	$
	we have $\nabla F(\omt;y) = 0$ or $\Kap(\Kap \omt - \yt) + \lambda \Kap \omt = 0$. Hence,
	$(\Kap + \lambda I) \Kap\omt = \Kap \yt$ or
	\begin{align}\label{eq:Kap:optim:cond}
	\Kap \omt = \Psil \yt, \quad \text{where}\;\; \Psil = (\Kap + \lambda I)^{-1} \Kap.
	\end{align}
	Let $w = (w_i) \in \reals^n$  be the noise vector in~\eqref{eq:mod:1} and $\wt = w/\sqrt{n}$.	
	We also let %
	\begin{align}\label{eq:us:def}
	\us = (\foms(x_i)) /\sqrt{n} = K \oms.
	\end{align}
	Then, we can write model~\eqref{eq:mod:1} as $\yt = \us + \wt$, where $\wt$ is zero mean with $\cov(\wt) = \sigma^2 I_n / n$.
	From~\eqref{eq:emp:uspace:equiv}, we have $\empnorm{\ft - \fs}^2 = \norm{K(\omt - \oms)}^2$, and
	\begin{align*}
	K(\omt - \oms) = \Kap \omt - \us = \Psil \yt - \us = (\Psil - I) \us + \Psil \wt,
	\end{align*}
	where the first equality uses assumption~\eqref{eq:omt:cond}.
	It follows that
	\begin{align*}
	\ex \vnorm{\ft - \fs}_n^2 = \vnorm{(I - \Psil) \us}^2 + \ex \vnorm{\Psil \wt}^2.
	\end{align*}
	where the first term is the approximation error (AE) and the second term, the estimation error (EE).
	Let us write $\Dap = \diag(\mu_1,\dots,\mu_r,0,\dots,0) \in \reals^{n \times n}$ so that $\Kap = U \Dap U^T$. We define
	\begin{align}\label{eq:Gaml:def}
	\Gaml = (\Dap + \lambda I)^{-1} \Dap, \quad \text{so that}, \quad \Psil = U \Gaml U^T
	\end{align}
	and note that $\Gaml$ is diagonal. Let $\vs = U^T \us$ and $\wtt = U^T \wt$. Then, since $\ell_2$ norm is unitarily invariant, we have 
	\begin{align*}
	\ex \vnorm{\ft - \fs}_n^2 = \vnorm{(I - \Gaml) \vs}^2 + \ex \vnorm{\Gaml \wtt}^2.
	\end{align*}

	\noindent \emph{Controlling the estimation error:} We have
	\begin{align*}
	\eerr := \ex\vnorm{\Gaml \wtt  }^2 = \ex \big[\wtt^T \Gaml^2 \wtt\big] 
	= \tr\big( \Gaml^2 \cov(\wtt)\big)
	= \frac{\sigma^2}{n} \tr(\Gaml^2),
	\end{align*}
	using $\cov(\wtt) = U^T \cov(\wt) U = (\sigma^2 / n) U^T U = \sigma^2 I_n / n$ since $U$ is an orthogonal matrix.
	Then,
	\begin{align}\label{eq:Gaml:diag}
	(\Gaml)_{ii} = \Big( \frac{\Dap_{ii}}{\Dap_{ii}+\lambda}\Big) = 
	\begin{cases}
	\frac{\mu_i}{\mu_i + \lambda}, & i =1,\dots,r \\
	0 & i=r+1,\dots,n,
	\end{cases}
	\end{align} 
	establishing the EE part of the result.

	\medskip
	\noindent \emph{Controlling the approximation error:} Recall that we are interested in the worst-case approximation error (WAE) over the unit ball of the Hilbert space, i.e., over $\fs \in \ball_\Hil$. Also, recall that without loss of generality, we can take $\fs = \foms$. Hence, 
	\begin{align}\label{eq:vs:ellips}
	1 \ge \hilnorm{f}^2 = \hilnorm{\foms}^2 = (\oms)^T K \oms = (\us)^T K^{-1} \us = (\vs)^T D^{-1} \vs
	\end{align}
	where the second equality is from~\eqref{eq:hilnorm:ident}, and the latter two are by definitions of $\us$ and $\vs = U^T \us$. We obtain
	\begin{align*}
	\waerr = 
	\sup_{(\vs)^T D^{-1} \vs \,\le\, 1} \vnorm{(I-\Gaml) \vs}^2.
	\end{align*}
	A further change of variable $\vs = D^{1/2} v$ gives
	\begin{align*}
	\waerr = \sup_{v^T v \, \le \, 1} \vnorm{(I-\Gaml) D^{1/2}v}^2 = \opnorm{(I-\Gaml) D^{1/2}}^2,
	\end{align*}
	where $\opnorm{\cdot}$, applied to matrices, is the $\ell_2\to \ell_2$ operator norm.
	Note that $\Gaml$ is a diagonal matrix with diagonal elements, $\mu_i/(\mu_i + \lambda)$ for $i=1,\dots,r$ followed by $n-r$ zeros. 
	It follows that $(I - \Gaml)D^{1/2}$ is diagonal with diagonal elements:
	\begin{align}\label{eq:Gamil:D:diag}
	[(I - \Gaml)D^{1/2}]_{ii}=
	\begin{cases}
	\frac{\lambda \sqrt{\mu_i}}{\lambda + \mu_i}, & i = 1,\dots,r,\\
	\sqrt{\mu_i}, & i = r+1,\dots,n.
	\end{cases}
	\end{align}
	Since $\{\mu_i\}$ is a non-increasing sequence, we obtain
	\begin{align*}
	\waerr = \max\Big\{ \max_{1 \le i \le r}\frac{\lambda^2 \mu_i}{(\lambda + \mu_i)^2}, \;
	\mu_{r+1} \Big\},
	\end{align*}
	which is the desired result. %
\end{proof}

\begin{proof}[Corollary~\ref{cor:domination}]
	Let $\eerr := \frac{\sigma^2}n \sum_{i=1}^r [\mu_i/(\mu_i+\lambda)]^2$ be the estimation error of $\ft_{r,\lambda}$ as in~\eqref{eq:krr:trunc:res}. Note that as long as $\mu_{r+1} > 0$, we have $\ee_{r,\lambda} < \ee_{r+1,\lambda} \le \ee_{n,\lambda}$. It remains to show that the WAE of the truncated KRR is less than that of full KRR. We have for $r \ge r(\lambda)$,
	\begin{align*}
	\wae_{r,\lambda} = \max\{H_r(\lambda),\mu_{r+1}\} \le \max\{H_n(\lambda),\mu_{r+1}\} = H_n(\lambda) = \wae_{n,\lambda}.
	\end{align*}
	This proves~\eqref{eq:max:risk:domination}. For the second assertion, it is enough to apply~\eqref{eq:max:risk:domination} with $\lambda = \lambda_n$, noting that in this case, the RHS will be the minimax risk of the full KRR and the LHS is further lower bounded by the minimax risk of the truncated KRR.
\end{proof}

\begin{proof}[of Corollary~\ref{cor:lambda:r}]
	Let us write $\wae_r(\lambda)$ and $E_r(\lambda)$ for the worst-case approximation and estimation errors, respectively, as a function of $\lambda$. Let $M_r(\lambda)$ be the worst-case MSE, so that $M_r(\lambda) = \wae_r(\lambda) + E_r(\lambda)$. The $\wae_r(\cdot)$ starts off with the constant branch $\wae_r(\lambda) = \mu_{r+1}$ for small values of $\lambda$. Let $h_i(\lambda) := h(\lambda; \mu_i)$. The constant branch starts at $\lambda = 0$ and extends to $\lambda =\lambda^{(1)}$ where $h_r(\lambda^{(1)}) = \mu_{r+1}$. Some algebra gives $\lambda^{(1)} = \mu_r / (\sqrt{\mu_r/\mu_{r+1}} - 1)$. For $\lambda \in [0, \lambda^{(1)}]$, we have $M_r'(\lambda) = E_r'(\lambda) < 0$ showing that the minimizer of $M_r$ is $\ge \lambda^{(1)}$.
	
	\begin{figure}[t]
		\centering
		\includegraphics[width=.5\textwidth]{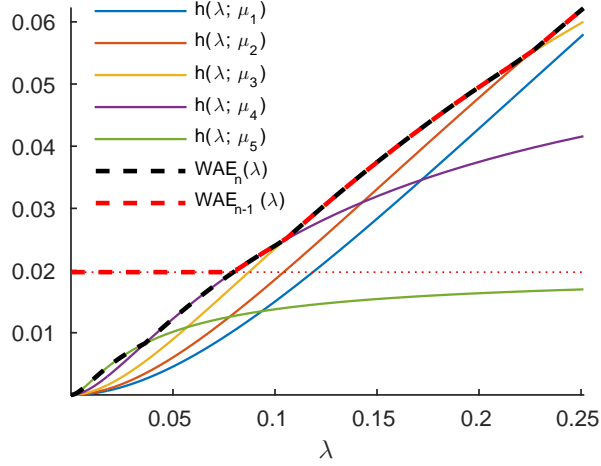}
		\caption{An illustration of $\lambda \mapsto \wae_r(\lambda)$ for a $5 \times 5$ Gaussian kernel matrix. The case $r=n=5$ corresponds to the full KRR and $r=n-1 = 4$ to a truncated version.}
		\label{fig:wae}
	\end{figure}
	The next branch of $\wae$ starts at $\lambda^{(1)}$ and ends at $\lambda^{(2)}$ which solves $h_r(\lambda^{(2)}) = h_{r-1}(\lambda^{(2)})$. The knots $\lambda^{(i)}$ determining subsequent branches are determined similarly:  $h_{r-i+2}(\lambda^{(i)}) = h_{r-i+1}(\lambda^{(i)})$ for $i=2,3,\dots,r$ and $\lambda^{(r+1)} = \infty$. We have $\wae_r(\lambda) = h_{r-i+1}(\lambda)$ for $\lambda \in I_i:= [\lambda^{(i)},\lambda^{(i+1)})$ for $i=1,\dots,r$. See Fig.~\ref{fig:wae}.
	
	Fix $i_* \in [r]$
	and let $j_* = r-i_*+1$. Then for $\lambda \in \text{int}(I_{i^*})$%
	\begin{align*}
	M_r'(\lambda) = h_{j_*}'(\lambda) + E'_r(\lambda) 
	&= \frac{2 \mu_{j^*}^2}{(\mu_{j^*} + \lambda)^3} \Big[ \Big( \lambda  - \frac{\sigma^2}{n} \Big) - \frac{\sigma^2}{n} \sum_{i\neq j_*} \frac{\mu_i^2}{(\mu_i + \lambda)^3} \frac{{(\mu_{j^*} + \lambda)^3}}{\mu_{j^*}^2}\Big]
	\end{align*}
	where $i$ ranges over $[r] \setminus\{j^*\}$.
	Note that
	\begin{align*}
	\sum_{i\neq j_*} \frac{\mu_i^2}{(\mu_i + \lambda)^3} \frac{{(\mu_{j^*} + \lambda)^3}}{\mu_{j^*}^2} \ge
	\Big( \sum_{i > j_*} \frac{\mu_{j^*}}{\mu_i}  + \sum_{i < j_*} \frac{\mu_i^2}{\mu_{j^*}^2} \Big)  \ge B_r.
	\end{align*}
	The first inequality is since $\lambda \mapsto (\mu_{j^*} + \lambda) / (\mu_i + \lambda)$ is increasing in $[0,\infty)$ if $\mu_{j^*} < \mu_i$, hence lower-bounded by its value at $\lambda = 0$, and is decreasing on $[0,\infty)$ if $\mu_{j^*} > \mu_i$, hence lower-bounded by its value as $\lambda \to \infty$. Then,
	\begin{align*}
	M'_r(\lambda) \le \frac{2 \mu_{j^*}^2}{(\mu_{j^*} + \lambda)^3} \Big[ \Big( \lambda  - \frac{\sigma^2}{n} \Big) - 
	\frac{\sigma^2}{n} B_r \Big) \Big].
	\end{align*}
	
	It follows that $M'_r(\lambda) < 0$ as long as $\lambda < \sigma^2 (1+B_r) /n$ no matter which interval $I_i$ contains $\lambda$. This shows that the minimizer of $M_r$ has to be $\ge \sigma^2 (1+B_r) /n$ completing the proof.
\end{proof}

\section*{Acknowledgement}
We thank Chad Hazlett and Linfan Zhang for helpful discussions and Zahra Razaee for comments on the manuscript.

\printbibliography

\appendix

\section{Time complexity comparison}\label{app:time}
The ST-KRR and approximate versions, such as Nystr\"{o}m and sketching, all have time complexity of $O( nr^2 +  r^3 ) = O(nr^2)$ for computing the $r$-truncated KRR estimate, once the pieces required for approximating the kernel matrix (e.g., $KS$ and $S^T K S$ in the case of sketching, $U_r$ and $D_r$ in the case of ST-KRR and so on) are computed. Computing these pieces is where these methods differ. For sketching, this step could have complexity as large as $O(n^2 r^2)$ for dense sketches, $O(n^2 \log r)$ for randomized Fourier and Hadamard sketches,  to as low as $O(nr)$ for the  Nystr\"{o}m. 

For the ST-KRR, this step involves computing the top-$r$ eigenpairs of the symmetric matrix $K$, for which the Lanczos algorithm is the standard and for which a complexity analysis is hard to find in the literature. However, results of~\cite{kuczynski1992estimating} suggest that it has average-case complexity $O(n^2 (r+\log n))$. More precisely,~\cite{kuczynski1992estimating} show that on average  $k = O( \log n /\sqrt{\eps})$ Lanczos iterations are enough to compute the top eigenvalue to  within relative error $\eps$, hence an overall average-case complexity $O( k N )$ where $N$ is the number of nonzero entries of matrix $K$.

\section{Remaining proofs}\label{app:rem:proofs}
\begin{proof}[Proposition~\ref{lem:approxerr:pluse:smoothness:good}]
	We will use the same notation as in the proof of Theorem~\ref{thm:krr:spec:trunc}. By the same argument as in that proof, we have $\empnorm{\fb - \fs}^2 = \vnorm{(I - \Gaml) \vs}^2$ where $\Gaml$ is defined in~\eqref{eq:Gaml:def} and $\vs = U^T \us$ for $\us$ given in~\eqref{eq:us:def}. Let $\omb$  be the solution of~\eqref{eq:omt:min} for the input $(\fs(x_i))$ (instead of $y$) so that $\fb = f_{\omb}$. Using the optimality condition in the proof of Theorem~\ref{thm:krr:spec:trunc}, 
	\begin{align*}
	\ub := K \omb = \Kt \omb = \Psil \us 
	\end{align*} 
	where we have used~\eqref{eq:omt:cond} and~\eqref{eq:Kap:optim:cond}, with $\yt = \us$ (i.e., $\wt = 0$).
	We can write
	\begin{align*}
	\hilnorm{\fb}^2 =  \omb^T K \omb = (\ub)^T K^{-1} \ub  
	= (\us)^T \Psil K^{-1} \Psil \us  
	= (\vs)^T \Gaml D^{-1} \Gaml \vs 
	\end{align*}
	using $\Psil = U \Gaml U^T$, $K = U D U^T$ and $\vs = U^T \us$. Recall from~\eqref{eq:vs:ellips} that $\fs \in \ball_\Hil$ is equivalent to $(\vs)^T D^{-1} \vs \le 1$. It follows that
	\begin{align*}
	\sup_{\fs \,\in\, \ball_\Hil}	\big(\, \vnorm{\fs - \fb}_n^2 + \lambda \vnorm{\fb}_\Hil^2 \,\big)
	&= \sup_{(\vs)^T D^{-1} \,\vs\, \le 1} \big[\,
	\vnorm{(I - \Gaml) \vs}^2 + \lambda  (\vs)^T \Gaml D^{-1} \Gaml \vs 
	\, \big] \\
	&= \sup_{(\vs)^T D^{-1} \,\vs\, \le 1} (\vs)^T \big[
	(I - \Gaml)^2 + \lambda  \, \Gaml D^{-1} \Gaml
	\big]  \vs \\
	&= \opnorm{D^{1/2} [	(I - \Gaml)^2 + \lambda  \, \Gaml D^{-1} \Gaml ] D^{1/2}} \\
	&= \opnorm{(I - \Gaml)^2 D + \lambda \Gaml^2}
	\end{align*}
	where the third equality is using the change of variable $\vs = D^{1/2}v$ as in the proof of Theorem~\ref{thm:krr:spec:trunc}, and the last line follows since all the matrices are diagonal and hence  commute. The result now follows by combining~\eqref{eq:Gaml:diag} and~\eqref{eq:Gamil:D:diag}, after some algebra.
\end{proof}

\begin{proof}[Corollary~\ref{cor:Gauss}]
	For any $a,b > 0$, we have
	$\frac12 (a \wedge b) \;\le\; (a^{-1} + b^{-1})^{-1} \;\le \;a \wedge b$, where $a \wedge b := \min\{a,b\}$. Hence, the estimation error in~\eqref{eq:krr:trunc:res} is bounded as
	\begin{align*}
	\eerr = \frac{\sigma^2}{n \lambda^2}  \sum_{i=1}^r \frac{\mu_i^2 \lambda^2}{(\mu_i + \lambda)^2} =
	\frac{\sigma^2}{n \lambda^2} \sum_{i=1}^r (\mu_i^{-1} + \lambda^{-1})^{-2} \le 
	\frac{\sigma^2}{n \lambda^2} \sum_{i=1}^r (\mu_i \wedge \lambda)^{2}.
	\end{align*}
	This upper bound is within a factor of $4$ of the estimation error.
	Using $\mu_i / (\lambda + \mu_i) \le 1$ to shave off the power by one, we obtain the weaker bound:
	\begin{align}
	\eerr = \frac{\sigma^2}{n \lambda}  \sum_{i=1}^r \frac{\mu_i^2 \lambda}{(\mu_i + \lambda)^2} \le
	\frac{\sigma^2}{n \lambda}  \sum_{i=1}^r \frac{\mu_i\lambda}{\mu_i + \lambda} \le
	\frac{\sigma^2}{n \lambda} \sum_{i=1}^r \mu_i \wedge \lambda. \label{eq:eerr:weak:bound}
	\end{align}
	Recalling  definition~\eqref{eq:Gauss:complexity}, we conclude that if $\lambda \ge \delta^2$, 
	\begin{align*}
	\eerr \le\frac{\sigma^2}{ n \delta^2} \sum_{i = 1}^r \min\{\mu_i,\delta^2\} = \Big( \frac{R_r(\delta)}{\delta} \Big)^2.
	\end{align*}
	Combining with the WAE bound~\eqref{eq:wae:bound}, we obtain the desired result.
\end{proof}

\begin{proof}[Lemma~\ref{lem:omega:u:equiv}]
	Let $F(\omega)$ and $G(u)$ be the objective functions in~\eqref{eq:krr:omg:1} and~\eqref{eq:krr:u:1}, respectively. We have $F(\omega) = G(K \omega)$ for any $\omega \in \reals^n$, which follows from the identity $K K^+ K = K$. Now, assume that $\omb$ is a minimizer of $F$, and let $\ub := K \omb$. Pick any $u \in \ran(K)$; there exists $\omega$ such that $u = K \omega$, and we have $G(\ub)  = F(\omb) \le F(\omega) = G(u)$. The other direction follows similarly.
\end{proof}

\section{Rate calculations}\label{app:rates}

Here we compute the error rate predicted by the strong and weak bounds and show that they are the same. Let $\gamma := \sigma^2/n$. Assume the polynomial eigendecay of the Sobolev-$\alpha$ kernel, i.e., $\mu_i \asymp i^{-2\alpha}$. Taking $k$ to be the smallest integer satisfying $k^{-2\alpha} \lesssim \delta^2$, we have
\begin{align*}
R_n^2(\delta) = \gamma \Big(k\delta^2 + \sum_{i = k+1}^n i^{-2\alpha}\Big) \le \gamma (k \delta^2 + k^{-2\alpha+1}) \lesssim \gamma k \delta^2
\end{align*}
where the first inequality uses an integral approximation to the sum and the second uses the definition of $k$. Setting $\delta^2 \asymp R_n(\delta)$ we have $\delta^2 \asymp \gamma k \asymp \gamma (\delta^2)^{-\frac1{2\alpha}}$, hence the critical radius $\delta_n^2 \asymp \gamma^{\frac{2\alpha}{\alpha+1}}$.

Now consider the strong bound. As discussed in the text, $\wae_{n,\lambda} \asymp \lambda$. Also, as the proof of Corollary~\ref{cor:Gauss} shows, we have 
\[
\ee_{n,\lambda} \asymp \frac{\gamma}{\lambda^2} \sum_{i=1}^n \min(\mu_i^2,\lambda^2).
\]
Letting $k$ be defined as the smallest integer such that $\mu_k \lesssim \lambda$, we  get $k^{-2\alpha} \lesssim \lambda$ as before. Then, the maximum MSE is bounded as
\begin{align*}
\mse \lesssim \lambda + \frac{\gamma}{\lambda^2} \Big(k \lambda^2 +  \sum_{i=k+1}^n i^{-4\alpha}\Big) \lesssim 
\lambda + \frac{\gamma}{\lambda^2} \Big(k \lambda^2 +  k^{-4\alpha+1}\Big).
\end{align*}
Since $k^{-4\alpha + 1} \lesssim k \lambda^2$, by the definition of $k$, we obtain $\mse \lesssim \lambda + \gamma k \lesssim \lambda + \gamma \lambda^{-\frac1{2\alpha}}$. Equating the two terms we obtain $\mse \asymp \lambda_n \asymp \gamma^{\frac{2\alpha}{2\alpha+1}}$ as before. 

For the Gaussian kernel, with $\mu_j \asymp e^{-c j \log j}$, it is not hard to verify that with $e^{-c k} \lesssim \lambda$, we get $\mse \lesssim \lambda + \gamma k \lesssim \lambda + \gamma \log(1/\lambda)$. Minimizing the bound over $\lambda$, we obtain $\lambda \asymp \gamma$, hence $\mse \lesssim \gamma \log (1/\gamma)$.

\end{document}